# LETTER TO SOUND RULES FOR ACCENTED LEXICON COMPRESSION

*Vincent Pagel[1], Kevin Lenzo[2] and Alan W. Black[3]*

[1] Faculté Polytechnique de Mons, Dolez 31, 7000 Mons, Belgium

[2] Carnegie Mellon University, 5000 Forbes Av, Pittsburgh PA15213, USA

[3] Centre for Speech Technology Research, University of Edinburgh, UK

cmp-lg/9808010  21 Aug 1998

## ABSTRACT

This paper presents trainable methods for generating letter to sound rules from a given lexicon for use in pronouncing out-of-vocabulary words and as a method for lexicon compression. As the relationship between a string of letters and a string of phonemes representing its pronunciation for many languages is not trivial, we discuss two alignment procedures, one fully automatic and one hand seeded which produce reasonable alignments of letters to phones (or epsilon). Top Down Induction Tree models are trained on the aligned entries. We show how combined phoneme/stress prediction is better than separate prediction processes, and still better when including in the model the last phonemes transcribed and part of speech information. For the lexicons we have tested, our models have a word accuracy (including stress) of 78% for OALD, 62% for CMU and 94% for BRULEX, allowing substantial reduction in the size of these lexicons.

## 1. MOTIVATION

In a text-to-speech (TTS) system, a major interest of building rule based grapheme-to-phoneme transcription systems is to treat out of vocabulary words (OOV). The secondary effect of storing rules is to reduce the memory amount required by the lexicon, which is of interest for hand-held devices such as talking dictionaries. The rule set can be viewed as a sort of compression algorithm that captures language regularities.

Those regularities are often disrupted by complex word morphology, and accentuation pattern in stress timed languages such as English or Dutch, which makes this field attractive for machine learning techniques.

Given a dictionary of words with stressed phonemic transcriptions, such as CMU [1], OALD [2] or BRULEX [3], one notices that when two word chunks have *similar* spelling they have a *similar* pronunciation. Two broad categories of algorithm can be used to learn those similarities:

- Grapheme to Phoneme (G2P) transducers treating variable graphic chunk sizes. Yvon gives a summary of available methods in [4]. Among which HMM where phonemes correspond to states emitting zero on more letters. Yvon also proposes an original chunk recombination method

- G2P with fixed size learning windows. A fixed set of attributes is comfortable for many learning techniques, which was initiated in the NetTalk system by Rosenberg et al.

Lazy learning techniques contribute to the second category with their amazing back-off abilities: if a test vector is not present in the training set, it is classified according to its most significant attributes (c.f. many papers by Daelemans and Bosch describing IGTREE [5]). The drawback of fixed size windows is that the word and its phonemic transcription have not the same length, hence one has to introduce empty symbols (noted epsilon) in the alphabet to align graphemic and phonemic representation and get a one to one correspondence

In the rest of this paper we deal with the second family of algorithms, and we propose solutions for both the grapheme/phoneme alignment and the grapheme-to-phoneme transcription.

## 2. GRAPHEME/PHONEME ALIGNMENT

The first problem to solve with alignment is that one letter can correspond to more than one phoneme, and that one phoneme can correspond to more than one letter. Since the learning technique we use requires a fixed size learning vector, one should introduce epsilons both in the graphemic and phonemic strings as in the example given table 1.

| **Graph**. | E | X | - | E | M | P | L | A | R | Y |
|---|---|---|---|---|---|---|---|---|---|---|
| **Phon**. | IH | G | Z | EH* | M | P | L | ER | - | IY |

**Table 1:** alignment of graphemes and stressed phonemes ( - stands for epsilon, and * is a primary accent)

For the languages we study in this paper (French and English) we can avoid introducing epsilons in the graphemic string since only few letters generate more than one phoneme. We define a short list of pseudo phonemes such as K_S or W_A (as found in the English word 'fax' and French word 'royal') so that all our corpora can have a one letter to one phoneme alignment.

Thus the alignment task becomes *"introduce epsilons in the phonemic representation so that it matches the length of the graphemic representation"*. We propose 2 solutions to solve this problem.

### 2.1 Automatic Epsilon Scattering Method

The idea is to estimate the probabilities for one grapheme G to match with one phoneme P, and to use DTW to introduce epsilons at positions maximizing the probability of the word's alignment path. Once the dictionary is aligned, the association probabilities can be computed again, and so on until convergence. Five such iterations have been found to be necessary on the CMU corpus.

Algorithm:
/* initialize prob(G ,P ) the probability of G matching P */
1. foreach word$_i$ in training_set
    count with DTW all **possible** G/P association for all possible epsilon positions in the phonetic transcription
/* EM loop */
2. foreach word$_i$ in training_set

$$alignment\_path = \arg\max \prod_{i,j} prob(G_i, P_j)$$
compute new_p(G,P) on alignment_path
3 if (prob != new_p ) goto 2

## 2.2 Hand-Seeded Method

The best alignment method we experimented with requires seeding with the set of feasible letter-phone (or pseudo phone) pairs regardless of context.

Thus for each letter a table is written of the possible phones it may match. E.g. "c" may go to /ch/, /s/, /k/, /sh/ or epsilon. For a given lexicon this is an easy incremental process to add to this table until most of the entries can be aligned. Once the table is built, all possible alignments for each entry are found.

The occurrences of each letter/phone pairs in these alignments are summed. A table of probabilities of phone given letter is estimated. Then all possible alignments are found again, but this time they are scored with respect to the probabilities of the letter/phone pairs. The most probable alignment is then selected.

In almost all cases this alignment appears to be that which a human labeler would select. Note that although this does require a human to seed the table no real expertise is needed, so this can be done even with only a little knowledge of the language the lexicon covers. As the table is built, entries that have no alignment are displayed, the number of which eventually reduces to a few per thousand. These are typically acronyms, abbreviations, foreign words etc, where there is in fact no obvious alignment and definitely not one that would be productive to learn.

## 2.3 The test corpora

In the rest of the paper we evaluate our algorithm on 3 dictionaries which are split into 90-10% partitions by selecting every tenth entry for the whole set for train and test procedure:

1. The Oxford Advanced Learner Dictionary (OALD) which contains 63399 British English entries including morphological variations, primary accents and Part of Speech.

2. The CMU release 0.6 contains 127070 American English entries, we use a 111726 subset of the CMU only containing entries that can be aligned. Notably this lexicon contains a substantial number of acronyms and proper names, many of which have a non-English origin.

3. The Brulex corpus contains 35743 French entries. Note that there is no flexion, which oversimplifies the task since it removes ambiguities between conjugated verbs/nouns, and ascertain the pronunciation of final 's' (many Latin words such as 'le bus').

## 2.4 Results

The result table 2 shows the difference in the accuracy of the models generated by the epsilon scattering method versus the hand-seeded method.

| Method | Word accuracy | Phone accuracy |
|---|---|---|
| *Epsilon scattering* | 63.97% | 90.69% |
| *Hand-seeded* | 78.13% | 93.97% |

**Table 2:** Performance on OALD vs. alignment method (using learning technique described below)

Ideally we would like to fully automatically extract the alignments and we see the hand seeded method as the target for our fully automatic method. We are still working on improving the epsilon scattering method.

## 3. LETTER TO PHONE TRANSDUCTION

Given a fixed size learning vector, we use Top Down Induction Trees to predict the corresponding phonemic output, epsilon being considered as a phoneme.

## 3.1 Learning technique

With ID3 [6], information gain is used to recursively determine which attribute in the learning vectors allow the best entropy gain between the full set and the partition of the set according to its attribute's values. The resulting structure is a decision tree containing questions and return values on terminal nodes. The difference with IGTREE [5] is that the information gain is not computed once for all for each attribute, but is computed again on each recursively split subset. This consumes small extra memory since the tree has to store which attribute is being tested, but allows different branches to test attributes in different orders. To rate compression ratio, we give tree sizes in the rest of this paper according to this formula:

> *size(tree)= if terminal_node(tree) then 1*
> *else 1 + /\* for the default return \*/*
> *sum foreach t (subtree(tree)) 1+size(t)*

To summarize, the size is the number of "*if*" tests in the tree plus the number of "*return*", which is directly proportional to the memory requirements (whether the tree is compiled as source code or downloaded in memory and interpreted).

Most of ID3 smoothing power lies in the *default* case statement, which returns the most probable value for a partial tree path according to the learning set. We have implemented the possibility not to develop branches when the information gain drops under a given threshold (over-training on the data).

We tried another implementation of decision trees (Wagon a CART [7] implementation) and the results we found were very similar. It appears that the alignment algorithm and vector content contributes much more to the accuracy of the models than the actual decision tree learning technique.

The classical learning vector is a graphemic sliding window that takes N letters on the left and N letters on the right of the letter being transcribed.

| Grapheme Vector | Phoneme + Stress |
|---|---|
| --- e xam | IH |
| --e x amp | G_Z |
| -ex a mpl | AE * |
| exa m ple | M |
| xam p le- | P |

**Table 3**: input vectors and corresponding P+S output, transcribing the beginning of the word "example" (note the use of the pseudo-phoneme G_Z)

## 4.1 Models For Stress Assignment

Some models previously treated the assignment of stress as a parallel task. In agreement with recent findings of Bosch et al. [5], we measured on the OALD corpus a drastic enhancement when merging phoneme and stress prediction in the same tree.

|  | Phoneme (no stress) | Phoneme + Stress | Word (no Stress) | Word+ Stress | Tree size |
|---|---|---|---|---|---|
| **2 Trees** | 95.6% | - | 73.1% | 54.6% | 24552 +103 |
| **Merged** | 95.4% | 94.8% | 69.4% | 69.3% | 30368 |

**Table 4**: tree for G2P + tree for stress **VS** single tree or G2PS

Table 4 shows the results on OALD corpus for comparing separate phone prediction followed by stress prediction as opposed to predicting both phones and stress with a single model. Although word accuracy for the models excluding stress and the stress model's accuracy (94.6%) are individually high their combined result (54.68%) is significantly lower than predicting the two together (69.36%). The Phoneme+Stress value is not available for the separate models, as the stress prediction model does not preserve phone alignment. Neither model currently has any explicit morphology, which is obviously relevant, as some stress cannot be assigned with just local context.

In the following we always include stress in our learning parameters (accented and unaccented vowels are considered as different phonemes).

## 4.2 Phonemic feedback

The transcription of a letter, with an N-sized context, is independent of the transcription job that has been carried out on the rest of the word before the current position. However hand-derived rule systems in French used to include phonetic context in their rules, to write more compact systems. The reason is that defining an open/closed syllable for example is straightforward with the right phonemic context, and tedious with graphemes. The phonetic feedback can also help balancing accents in the word.

**Left to Right or Right to Left.** Of course one can only introduce the phonemes that have already been transcribed, thus if one needs the phonemes on the right one must transcribe the letters from left to right, and vice versa.

| Corpus type | Word +Stress | Phoneme +Stress | Tree size |
|---|---|---|---|
| **OALD 3 letters** | 73.41 | 92.74 | 57395 |
| **OALD 3 letters + 3 last phonemes L to R** | 75.46 | 92.62 | 59667 |
| **OALD 3 letters + 3 last phonemes R to L** | 76.66 | 93.60 | 56299 |
| **BRULEX 3 letters** | 93.74 | 97.76 | 9917 |
| **BRULEX 3 letters + 3 last phonemes L to R** | 94.05 | 98.23 | 8743 |
| **BRULEX 3 letters + 3 last phonemes R to L** | 94.34 | 98.84 | 9059 |
| **CMU 3 letters** | 59.71 | 86.95 | 127393 |
| **CMU 3 letters + 3 last phonemes, L to R** | 62.79 | 87.84 | 123301 |
| **CMU 3 letters + 3 last phonemes, R to L** | 61.40 | 87.90 | 118767 |

**Table 5**: tree performance and size when including in the context vectors the 3 last phoneme transcribed (depends on the transcription direction, Left to Right, or Right to Left)

**Evaluation.** The enhancement shown table 5 for French is marginal, which means that the tree was already embedding syllable information derived from the letter sequence (what is tedious for a human being need not be for a decision tree). However the phonemic feedback clearly simplifies the decision tree (12% smaller).

English corpora benefit from both tree simplification and performance enhancement. The advantage of the right to left transcription direction can be explained by the fact that most of the time the end of the word gives indication on its morphology (hence on its accentuation pattern).

For example stress shifts like in *'strategy' (S T R AE\* T AH JH IY) / 'strategic' (S T R AH T IY\* JH IH K)* cannot be handled by a system provided with a 3 letter context and left to right transcription. As a matter of fact, when transcribing the 'a' with information 'str a teg' one cannot decide between AE* or AH.

On the other hand, with a right to left transcription, the information vector is either 'str a teg' + T IY* JH, either 'str a teg' + T AH JH. Two successive syllables can't be accentuated; the system has thus enough information to correctly decide between the 2 options.

### 4.3 Including Part Of Speech

Heterophonic-homographs are quite common in English and French, and can be disambiguated when their part of speech is known (many verb-noun or verb-adjective pairs).

| Corpus type | Word+ Stress | Phoneme+ stress | Tree size |
|---|---|---|---|
| **OALD 3 letters** | 73.41% | 92.74% | 57395 |
| **OALD 3 letters + Part of Speech** | 75.73% | 93.19% | 61671 |
| **OALD 3 leters + Part of Speech + 3 phones left to right** | 78.13% | 93.97% | 59135 |

**Table 6:** Influence of POS alone, and POS + 3 last phonemes.

Including a POS tag in each learning vector to indicate the nature of the word to which it belongs is easy, this enhance the word accuracy by 2.3% as shown table 6. The synergy of POS with the phonemic feedback in a left to right transcription is excellent, as the resulting gain is nearly the sum of their gain independently.

## 4. APPLICATION TO COMPRESSION

All the results given from the start of the paper are results about generalization performances. Are those methods applicable for the compression of dictionaries? To evaluate the compression, let's recall that the training and testing sets of the G2P model are the same.

Depending on the amount of memory occupied by a node in the tree, and on the size of the exception lexicon, the developer can choose a memory trade-off by diminishing the depth of the decision tree as shown figure 1. On OALD for example, the best compression result is 61831 nodes to represent 99.02% of the corpus (that is an exception lexicon of 621 entries, and a compression ratio of 1 to 22 for the text version of OALD).

## 5. CONCLUSION

This paper has presented a method building letter to sound rules for a given lexicon in a general language independent way. We have tested it on English and French and feel it suitable for many other languages.

From our results it seems that over-training is generally not a problem, more data for the rules is always useful. However this may be partly be due to the way we selected our test sets (one entry out of ten). As the lexicon is in alphabetical order it is likely that the words that were immediately next to test entries are very similar. The question of accuracy with respect to genuinely unknown words is discussed more fully in [8].

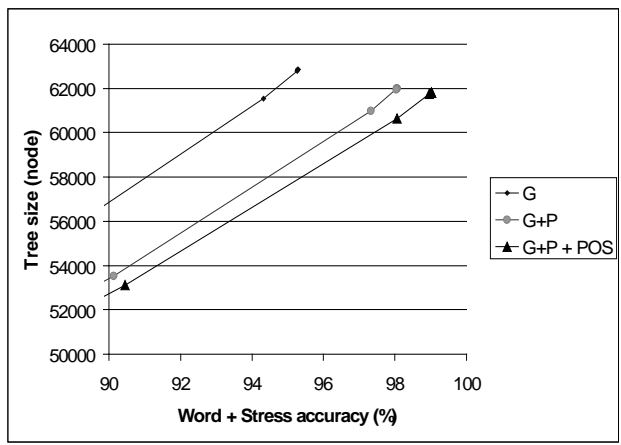

**Figure 1**: OALD percentage of correct word (train/test sets are the same) as a function of the tree size for Grapheme only, Grapheme+Phoneme, Grapheme+Phoneme+Part of Speech

The automatic learning programs described in this paper as well as speaking dictionaries are available from the MBROLA project [9] home page http://tcts.fpms.ac.be/synthesis/mbrdico

## 6. REFERENCES

1. Weide R. L, "Carnegie Mellon Pronouncing Dictionary" release 0.6 , www.cs.cmu.edu , 1998

2. Mitten R. "Computer-usable version of Oxford Advanced Learner's Dictionary of Current English" Oxford Text Archive, 1992.

3. Content A., Mousty P. and Radeau M. "BRULEX: Une base de données lexicales informatisée pour le français écrit et parlé", *L'Année Psychologique*, 1990, p551-566

4. Yvon F. "Prononcer par analogie: motivation formalisation et évaluation", Phd thesis, ENST, 1996

5. Van den Bosch A., Weijters T and Daelemans W. "Modularity in inductive-learned word pronunciation systems" in proc. NeMLaP3/CoNNL98 / Powers D.M.W., Sydney, 1998, p. 185-194

6. Quinlan J. R., "*C4.5 Programs for Machine Learning*" San Mateo, CA Morgan Kaufman, 1993

7. Breiman L., Friedman J., Olshen R. and Stone C., "Classification and Regression Trees", Wadsworth & Brooks, Pacific Grove, CA., 1984

8. Black A.W., Lenzo K., Pagel V. "Issues in Building General Letter to Sound Rules", ESCA Synthesis Workshop, Australia 1998

9. Dutoit T., Pagel V., Pierret N., Bataille F., Van der Vrecken O. "The MBROLA Project: Towards a Set of High-Quality Speech Synthesizers Free of Use for Non-Commercial Purpose", proc ICSLP'96,vol.3, p1393-1396